\documentclass[letterpaper]{article} 
\usepackage{aaai24}  
\usepackage{times}  
\usepackage{helvet}  
\usepackage{courier}  
\usepackage[hyphens]{url}  
\usepackage{graphicx} 
\usepackage{natbib}  
\usepackage{caption} 
\usepackage{algorithm}
\usepackage{algorithmic}
\usepackage{amsmath}
\usepackage{enumitem}
\title{T-REX: Vision-Based System for Autonomous Leaf Detection and Grasp Estimation}
\author{
    Srecharan Selvam\textsuperscript{\rm 1},
    Abhisesh Silwal\textsuperscript{\rm 1},
    George Kantor\textsuperscript{\rm 1}
}
\affiliations{
    \textsuperscript{\rm 1}Robotics Institute, Carnegie Mellon University, Pittsburgh, PA, 15213\\
    sselvam@andrew.cmu.edu
}

\begin{document}

\maketitle

\begin{abstract}
T-Rex (The Robot for Extracting Leaf Samples) is a gantry-based robotic system developed for autonomous leaf localization, selection, and grasping in greenhouse environments. The system integrates a 6-degree-of-freedom manipulator with a stereo vision pipeline to identify and interact with target leaves. YOLOv8 is used for real-time leaf segmentation, and RAFT-Stereo provides dense depth maps, allowing the reconstruction of 3D leaf masks. These observations are processed through a leaf grasping algorithm that selects the optimal leaf based on clutter, visibility, and distance, and determines a grasp point by analyzing local surface flatness, top-down approachability, and margin from edges. The selected grasp point guides a trajectory executed by ROS-based motion controllers, driving a custom microneedle-equipped end-effector to clamp the leaf and simulate tissue sampling. Experiments conducted with artificial plants under varied poses demonstrate that the T-Rex system can consistently detect, plan, and perform physical interactions with plant-like targets, achieving a grasp success rate of 66.6\%. This paper presents the system architecture, implementation, and testing of T-Rex as a step toward plant sampling automation in Controlled Environment Agriculture (CEA).
\end{abstract}

\section{Introduction}

Plant diseases pose a persistent threat to global agricultural productivity, with estimated yield losses ranging from 20--40\% annually due to pathogenic infestations and pest attacks \cite{Savary2012, Savary2019}. These losses have far-reaching implications—not only impacting food security but also the economic viability of growers. In many cases, outbreaks are exacerbated by the delayed detection of disease, where symptoms only manifest visibly at an advanced stage of infection. Traditional practices for identifying plant pathogens—such as visual inspection, culturing of microbes, and PCR-based molecular confirmation—are slow, manual, and not scalable for large production systems \cite{Savary2012, Pilli2014}. These approaches often take several days to weeks from symptom detection to actionable diagnosis, allowing the disease to spread unnoticed across farms and greenhouses.

Recent advancements in artificial intelligence (AI), robotics, and molecular biology have opened new frontiers in precision agriculture. In particular, AI-powered computer vision systems are capable of identifying early stress indicators in plants based on image-based features, even before visual symptoms are apparent to the human eye \cite{Chin2023}. Robotic platforms, including drones and ground rovers, have been used to scout large agricultural areas autonomously and collect image data for disease diagnosis \cite{Johnson2023, Trippa2024}. At the same time, genomic technologies—especially portable, real-time DNA sequencing platforms like Oxford Nanopore—have enabled high-resolution identification of plant pathogens directly from leaf tissue, eliminating the need for lengthy culture-based assays \cite{Espindola2015}. When combined, these technologies form the backbone of an emerging paradigm in agriculture: cyber-physical systems (CPS) for closed-loop plant health monitoring and control.

Despite these promising developments, significant limitations remain. Many of the current AI-based disease detection systems rely solely on imaging, which can produce false positives/negatives when differentiating between abiotic stress (e.g., drought) and biotic infections (e.g., bacterial or fungal pathogens) \cite{Johnson2023}. On the other hand, while DNA sequencing offers definitive diagnosis, it still depends on manual tissue sampling—a process that is both time-consuming and labor-intensive. Furthermore, few systems have successfully integrated real-time imaging, autonomous sampling, and in-field sequencing into a cohesive, closed-loop disease control workflow.

Controlled Environment Agriculture (CEA)—including greenhouses, vertical farms, and hydroponic systems—has emerged as a promising solution for high-yield, resource-efficient crop production. These systems allow year-round cultivation under tightly regulated environmental conditions and are increasingly adopted in urban and climate-sensitive regions \cite{Patel2021}. However, CEA systems also face critical challenges: high energy demands for lighting and HVAC, limited scalability, and the need for skilled labor to manage complex microclimates \cite{Entech2023, CEAgWorld2024}. These limitations underscore the importance of automation and robotic solutions to improve reliability and reduce manual intervention in such controlled setups.

This paper focuses on the development and implementation of T-Rex, a gantry-type robotic system designed for autonomous detection and diagnosis of plant diseases through leaf grasping and DNA extraction. T-Rex (Robot for EXtracting leaf samples) operates within a high-density greenhouse setup, targeting crops such as tomato and soybean, and is capable of executing an entire disease detection pipeline—from plant scanning and leaf analysis to sampling and sequencing—without human intervention \cite{Silwal2024-AAAI}.

The imaging component of T-Rex uses stereo cameras and the YOLOv8 segmentation model to detect early disease symptoms such as wilting, discoloration, or stunted growth \cite{Trippa2024}. These cues are used to guide a six-degree-of-freedom (6-DOF) gantry manipulator, which physically interacts with selected leaves using a novel microneedle-based end-effector \cite{Silwal2024-AAAI}. This end-effector clamps the leaf with minimal deformation and extracts tissue samples by piercing it with microneedle arrays. The extracted tissue is immediately transferred to a microfluidic sequencing pipeline, where low-volume samples undergo library preparation for real-time nanopore sequencing \cite{Espindola2015}. This enables the system to not only detect symptoms but also confirm the presence of pathogens using genetic evidence, all within the greenhouse environment.

\section{System Overview}

\subsection{System Design}

\textbf{Mechanical Design:} The T-Rex robot is built as a gantry system spanning a 3 m $\times$ 1.5 m growing area \cite{Silwal2024-AAAI}. This design provides coverage of multiple plants arranged on a flat bench, similar to modern hydroponic greenhouse layouts. The robot's kinematics consist of three prismatic axes (for $X$, $Y$, $Z$ positioning) and three revolute joints on the end-effector, denoted 3P3R \cite{Silwal2024-AAAI}. Together, these achieve a full 6-DOF manipulation workspace, allowing the end-effector to be positioned and oriented anywhere within the plant canopy volume. The prismatic axes move a carriage in the horizontal plane and raise/lower the end-effector, while the wrist joints adjust pitch, yaw, and roll of the sampling tool. This enables, for instance, approaching a leaf from directly above or at slight angles if needed. All actuators are high-torque digital servo motors controlled in firmware, offering positional accuracy sufficient for repeatable fine motions \cite{Silwal2024-AAAI}.\\

\begin{figure*}[t]  
    \centering
    \includegraphics[width=0.95\textwidth]{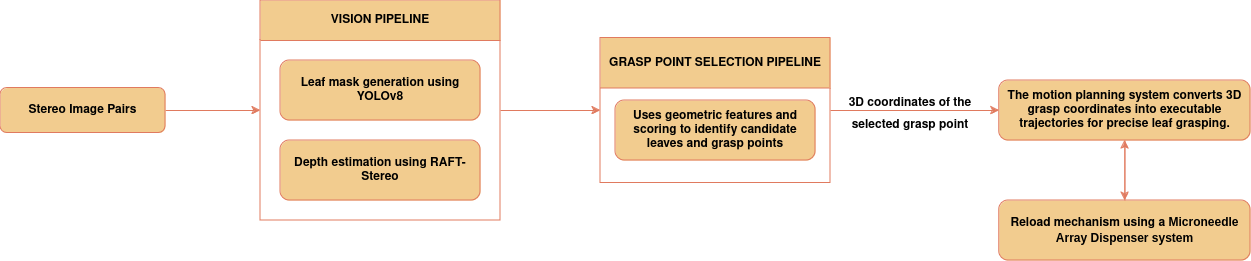}  
    \caption{T-REX system architecture showing the main processing pipeline: initial stereo image acquisition, perception and leaf grasping analysis, motion planning for execution, leaf grasping operations, and the reload mechanism for continuous operation. The feedback loops (dotted lines) enable autonomous multi-sample collection without human intervention.}
    \label{fig:trex-system-arch}
\end{figure*}

\begin{figure}[H]
    \centering
    \includegraphics[width=0.99\linewidth]{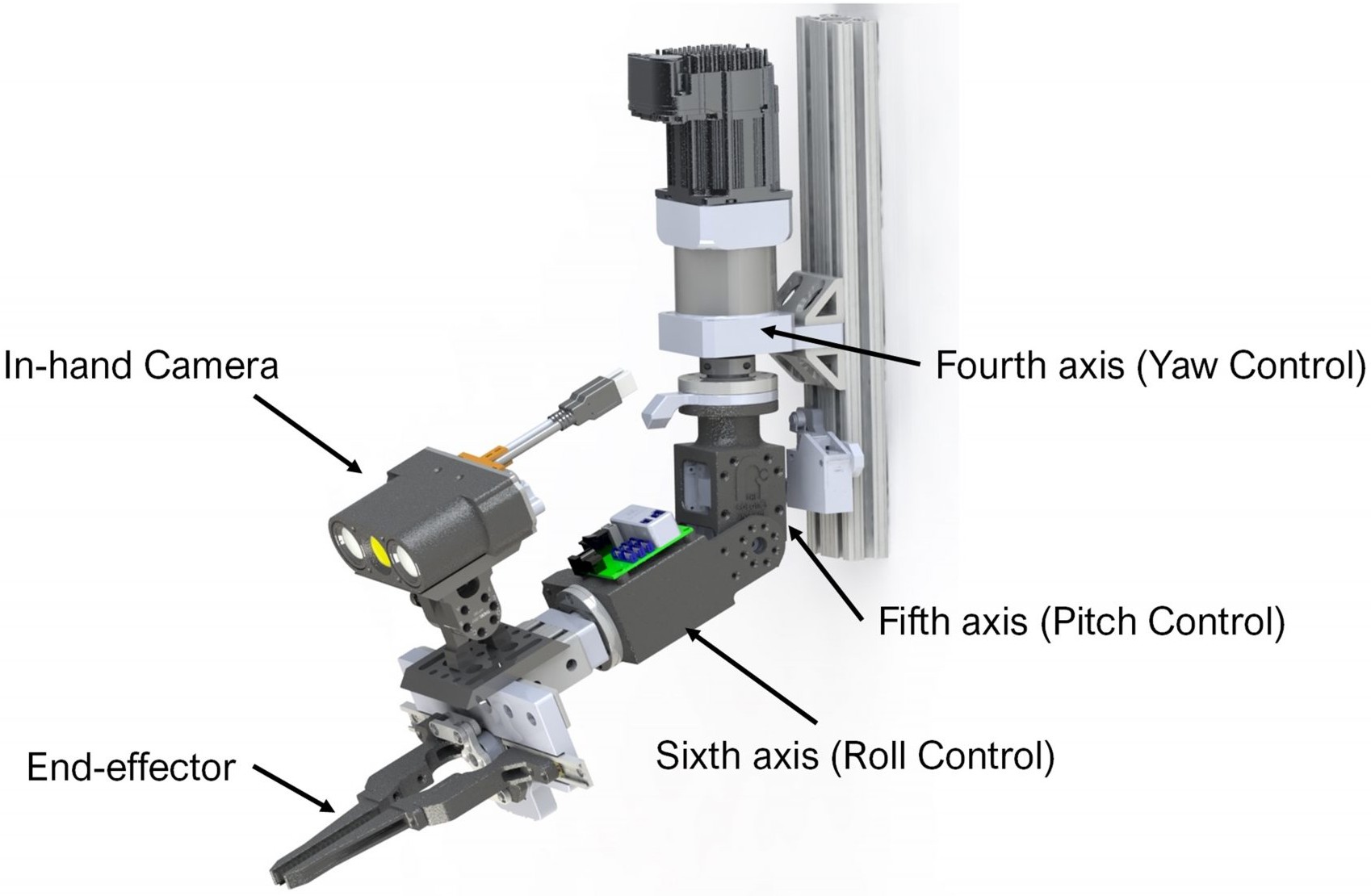}
    \caption{CAD rendering of T-Rex’s wrist and end-effector subsystem. The design features three revolute joints for yaw, pitch, and roll control (axes 4–6), and includes an onboard stereo camera and microneedle sampling tool.}
    \label{fig:end-effector-cad}
\end{figure}

\textbf{End-Effector:} The end-effector is a critical component of the T-Rex system, designed for precise leaf sampling (Figure \ref{fig:end-effector-cad}). It consists of two side grippers that move laterally (left and right) controlled by a Dynamixel motor. When activated, these side components move toward each other to grasp the target leaf. The end-effector also incorporates a stepper motor that performs vertical (up and down) linear motion. After the side grippers secure the leaf, the stepper motor moves downward to initiate the sampling process, pressing the microneedle array against the leaf surface.

The microneedle array is housed in a thin plastic holder with laser-etched breakaway tabs. When pressed against the leaf, the needles penetrate the leaf surface to extract sap samples \cite{Silwal2024-AAAI}. After sampling, the stepper motor returns to its upper position, and the side grippers move apart to release the leaf. The system includes an automated reloading mechanism that uses magnetic transfer to precisely position new microneedle arrays while ejecting used ones. This magnetic design features strategically placed poles (attractive at one end, repulsive at the other) to facilitate smooth transitions between arrays without requiring complex force control.\\

\begin{figure}[H]
    \centering
    \includegraphics[width=0.95\linewidth]{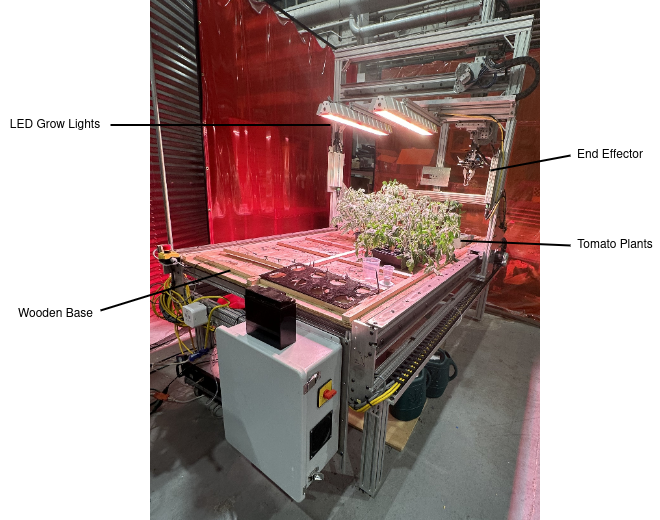}
    \caption{The T-Rex gantry robot setup inside a controlled lab environment. It spans a 3m $\times$ 1.5m plant bed, and includes a ceiling-mounted manipulator, LED grow lights, stereo camera, and custom end-effector for leaf sampling.}
    \label{fig:system_architecture}
\end{figure}

\textbf{Sensing and Control Architecture:} An active-stereo camera is mounted on the end-effector to capture stereo RGB images of the plants below. The camera includes an active light source for consistent illumination \cite{Silwal2021}, important in greenhouse environments with variable lighting. The robot operates under the Robot Operating System framework \cite{Silwal2024-AAAI}, which organizes the software into distributed nodes for perception, planning, and actuation. A centralized master node coordinates the pipeline: it triggers image capture, receives processed vision data, decides on a target leaf/grasp, and commands the motion controllers. The vision pipeline leverages pre-trained deep models for robust perception, with YOLOv8 achieving segmentation at 0.15 seconds latency and RAFT-Stereo producing depth maps at 0.55 seconds per frame, enabling near real-time operation. The gantry's positional feedback comes from encoders on each axis, providing sub-millimeter positioning accuracy essential for precise microneedle insertion. Low-level joint control is handled by ROS control packages (trajectory controllers) with real-time loops for smooth motion \cite{Macenski2022}. The overall system setup (shown in Figure \ref{fig:system_architecture}) follows a modular design: a \textit{YOLOv8 segmentation node} identifies leaves, a \textit{RAFT-Stereo node} computes depth, a \textit{grasp planning module} selects targets, and a \textit{motion executor} carries out the gantry movements \cite{Silwal2024-AAAI}. The system architecture also incorporates safety mechanisms including emergency stops, workspace limit enforcement, and collision detection through force-torque sensing on the Z-axis. This modular approach also allows easy swapping or upgrading of components (for example, replacing the vision module with a new algorithm).
  
\subsection{Vision Pipeline}

\begin{figure}[t]
    \centering
    \includegraphics[width=\columnwidth]{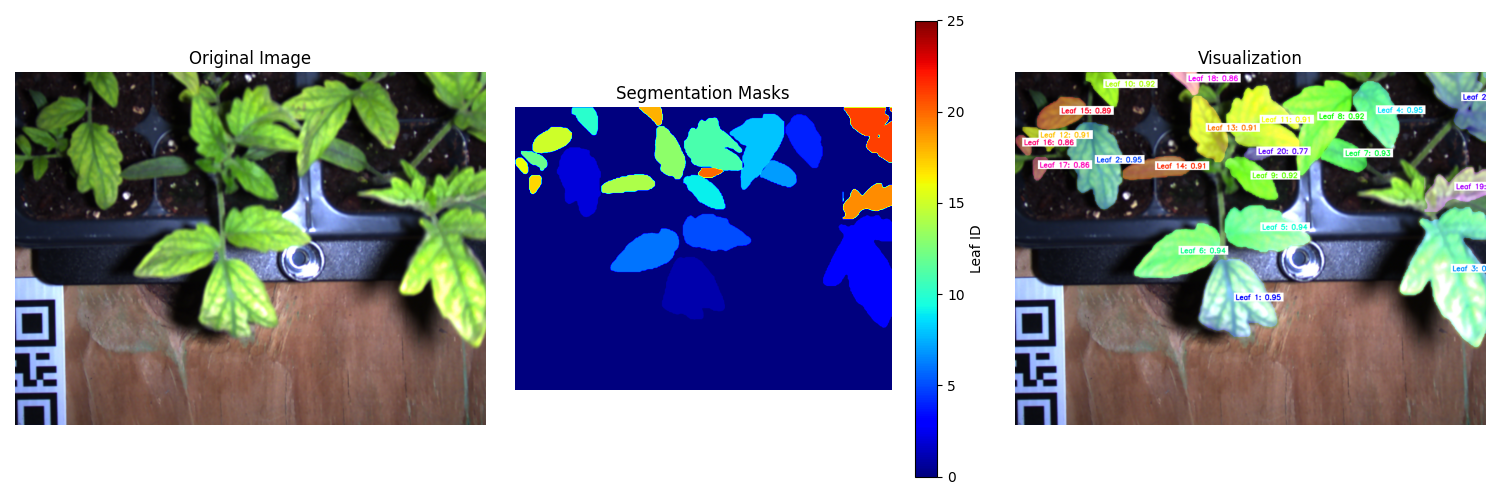}
    \caption{YOLOv8 leaf segmentation pipeline. Left: Original top-down image of a tomato plant tray. Center: Segmentation mask output, where each color represents a unique leaf ID. Right: Visualization overlay showing confidence scores assigned to each detected leaf.}
    \label{fig:yolo-pipeline}
\end{figure}

\begin{figure*}[t]
    \centering
    \includegraphics[width=0.95\textwidth]{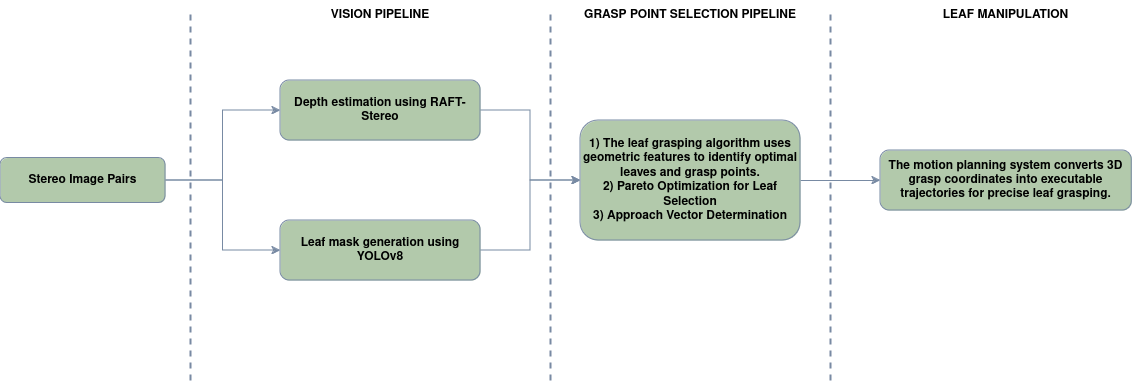}
    \caption{T-Rex system block diagram showing the complete pipeline from stereo image acquisition through vision processing, grasp point selection, motion planning, and reload mechanism.}
    \label{fig:system-block-diagram}
\end{figure*}

\textbf{YOLOv8 Segmentation:} Accurate perception of individual leaves is central to our approach. T-Rex's vision system uses a multi-stage pipeline to transform raw stereo images into rich 3D information about each leaf \cite{Silwal2024-AAAI}, as illustrated in the system overview (Figure~\ref{fig:system-block-diagram}). First, an instance segmentation model identifies all leaves in the camera's view. We implemented this using Ultralytics YOLOv8, a one-stage object detector with built-in segmentation capabilities \cite{Trippa2024}. The network was fine-tuned on a custom dataset of top-down images of tomato and soybean canopies. Each image was annotated for "leaf" instances (Figure~\ref{fig:yolo-pipeline}). During operation, YOLOv8 runs on each frame ($\sim$1440$\times$1080 resolution) and outputs a set of leaf masks, each with a confidence score >0.9 on average. Post-processing merges masks across the stereo pair and assigns a unique ID to each leaf, enabling multi-leaf tracking if the camera moves \cite{Silwal2024-AAAI}. The result is a 2D map of segmented leaves in the image.\\

\textbf{RAFT-Stereo Depth Estimation:} Next, to recover depth, we utilize RAFT-Stereo, a deep learning model for dense stereo matching. RAFT-Stereo (a variant of the RAFT optical flow network by Teed and Deng) computes a high-resolution disparity map from the synchronized left and right camera images \cite{Teed2020, Lipson2021}. It uses a recurrent neural network to iteratively refine pixel correspondences and achieves sub-pixel disparity accuracy on 1080p image pairs \cite{Lipson2021}. Our depth estimation pipeline completes in $\sim$150 ms per stereo pair \cite{Lipson2021}, fast enough for near-real-time analysis. The output is a dense depth image in which each pixel's value corresponds to its distance from the camera. We calibrate the stereo camera rigorously beforehand (intrinsics and extrinsics), so these depth values are accurate to within a few millimeters in the working range.\\

\begin{figure}[t]
    \centering
    \begin{minipage}{0.49\columnwidth}
        \centering
        \includegraphics[width=\textwidth]{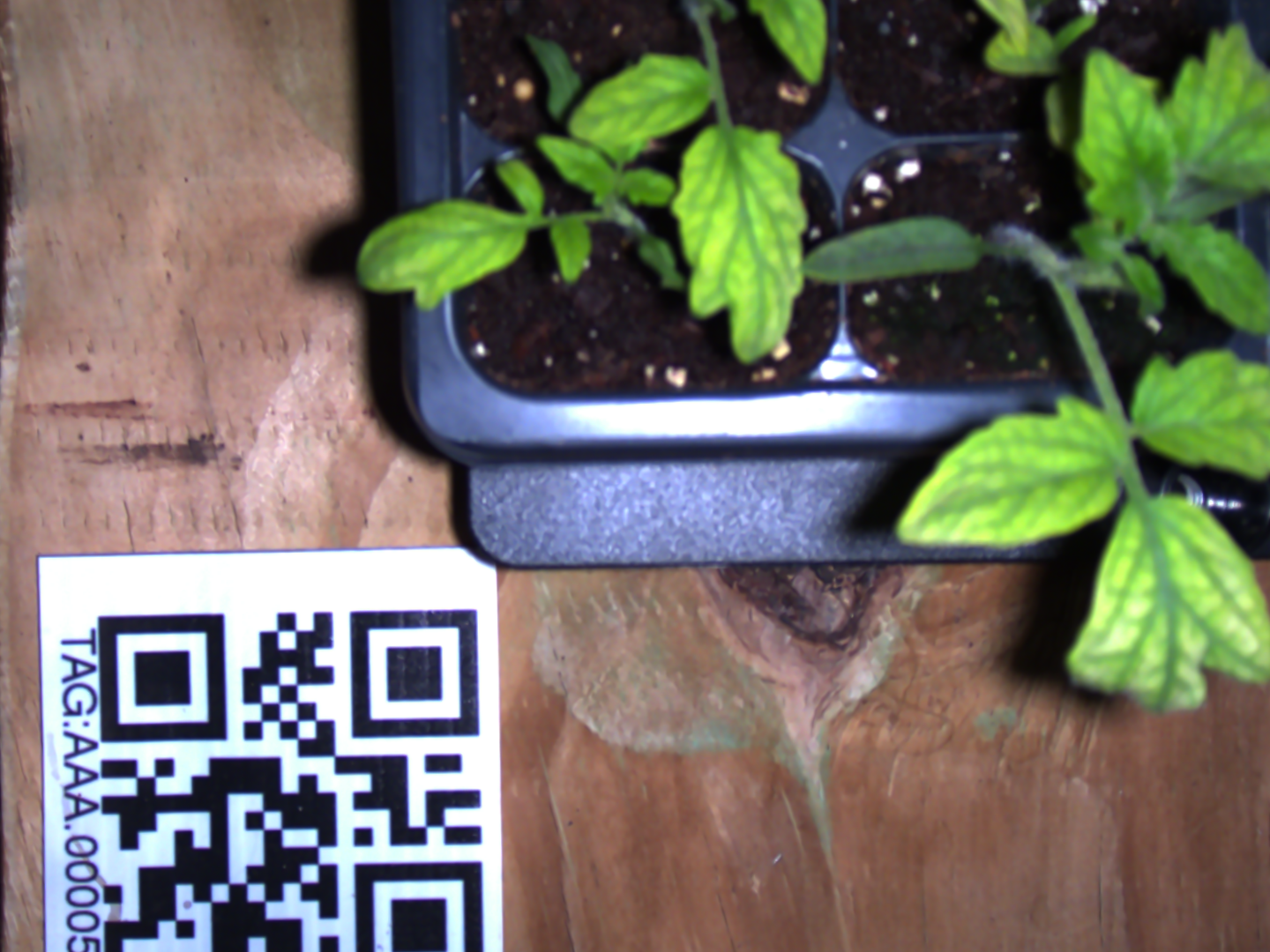}
        \vspace{-0.3cm}
        \caption*{(a) Raw stereo image}
    \end{minipage}%
    \hfill
    \begin{minipage}{0.49\columnwidth}
        \centering
        \includegraphics[width=\textwidth]{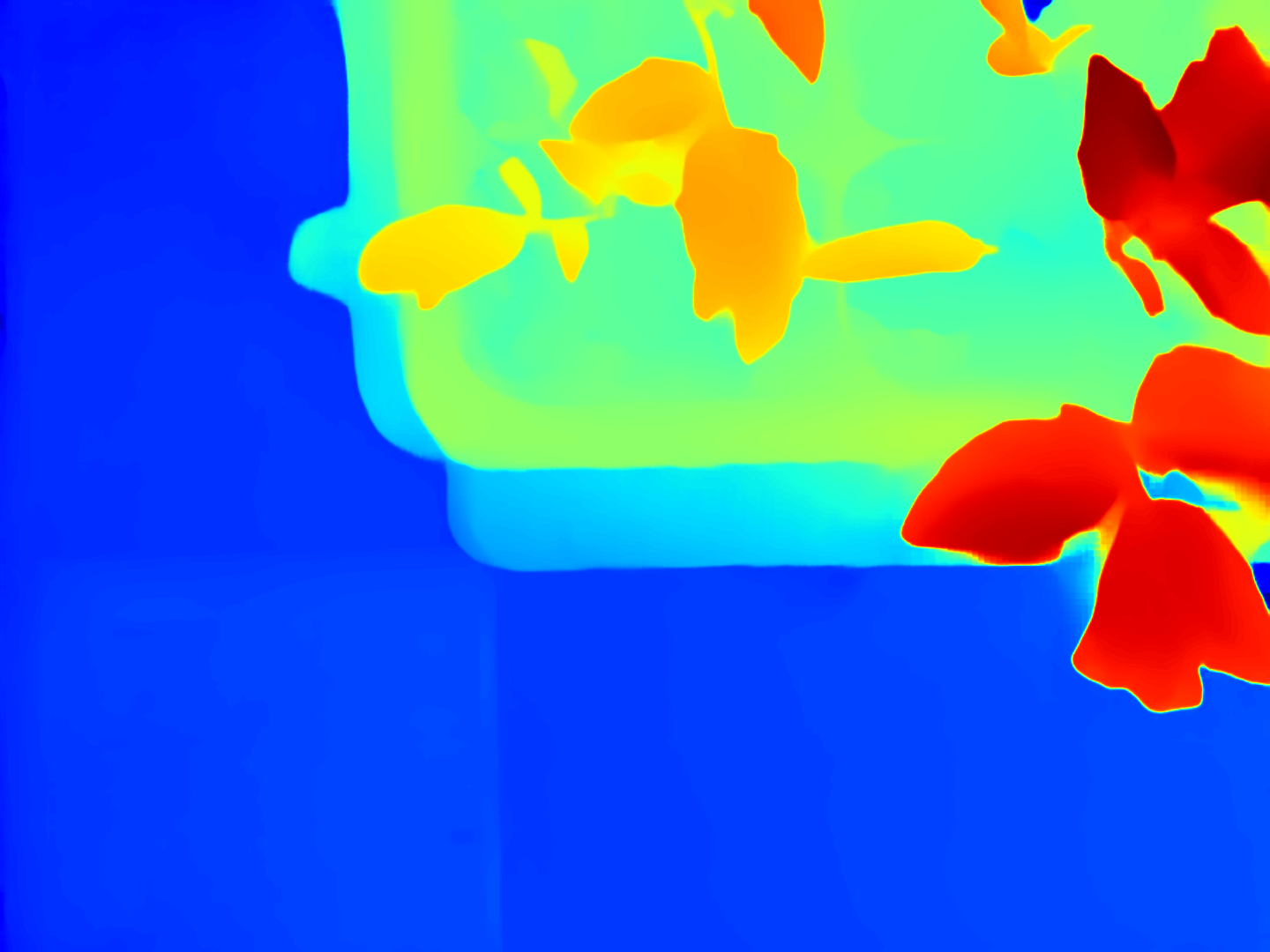}
        \vspace{-0.3cm}
        \caption*{(b) RAFT-Stereo depth map}
    \end{minipage}
    \\[0.2cm]
    \begin{minipage}{0.82\columnwidth}
        \centering
        \includegraphics[width=\textwidth]{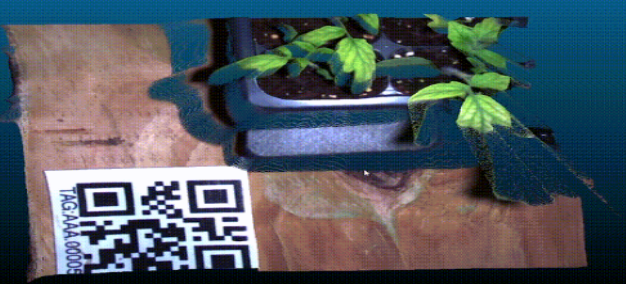}
        \vspace{-0.3cm}
        \caption*{(c) 3D leaf reconstruction from stereo depth and segmentation}
    \end{minipage}
    \caption{RAFT-Stereo outputs showing the processing pipeline: raw image, depth map, and 3D reconstruction.}
    \label{fig:raft-stereo-outputs}
\end{figure}

\begin{figure}[t]
    \centering
    \begin{minipage}{0.49\columnwidth}
        \centering
        \includegraphics[width=\textwidth]{figure/raw_image.png}
        \vspace{-0.3cm}
        \caption*{(a) Raw image of the plant with leaf candidates}
    \end{minipage}%
    \hfill
    \begin{minipage}{0.49\columnwidth}
        \centering
        \includegraphics[width=\textwidth]{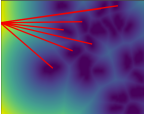}
        \vspace{-0.3cm}
        \caption*{(b) Signed Distance Field (SDF) representation used for grasp planning}
    \end{minipage}
    \caption{Visual comparison of the raw plant image and its corresponding SDF representation used for grasp planning. The SDF visualization shows free space (blue) and occupied regions (yellow/red). Red rays illustrate potential grasp planning directions.}
    \label{fig:sdf-heatmap}
\end{figure}

\textbf{3D Leaf Reconstruction:} Finally, the 2D segmentation and disparity are combined to produce 3D leaf information. For each segmented leaf mask, the corresponding depth pixels are extracted and reprojected into 3D space using the camera's calibration parameters. This yields a sparse point cloud for each leaf. By aggregating these points, we compute properties like the leaf's centroid in 3D, its approximate surface normal, and area. Essentially, we obtain a 3D-aware mask for each leaf—the system knows which points in space belong to which leaf. Figure~\ref{fig:raft-stereo-outputs} shows an example of the resulting 3D point cloud, reconstructed by combining the depth map with the YOLOv8-generated leaf masks (Figure~\ref{fig:yolo-pipeline}). To support downstream motion planning, a Signed Distance Field (SDF) is computed from this 3D data to model occupied and free space. Figure~\ref{fig:sdf-heatmap} visualizes the SDF, where cooler colors indicate navigable free space and warmer colors represent plant-obstructed regions. The red rays indicate candidate grasp approach directions explored during planning. This rich perception output forms the basis for intelligent grasp point selection and trajectory planning in the next stage. Notably, the entire vision pipeline leverages pre-trained deep models (YOLOv8 and RAFT-Stereo) for robust perception but focuses our contributions on the 3D geometric reasoning and decision logic that follow.

\subsection{Grasp Point Selection Pipeline}

Given the set of segmented 3D leaves from the vision module, T-Rex must decide which leaf to sample and where to grasp it. We formulate this as a two-step process: (1) \textit{Optimal leaf selection} among all detected leaves, and (2) \textit{Optimal grasp point selection} on that chosen leaf. Our approach relies on geometric heuristics (hand-crafted CV algorithms) rather than learned models, to ensure interpretability and avoid the need for labeled grasp data. Below we describe the criteria and scoring system used in each step.

\begin{equation}
\begin{aligned}
L^* = \underset{L_i \in \mathcal{L}}{\arg\max} \Big( &w_c S_c(L_i) \\
&+ w_d S_d(L_i) + w_v S_v(L_i) \Big)
\end{aligned}
\end{equation}

Where:
\begin{itemize}[label={--}]
    \item $L^*$ is the optimal leaf selection
    \item $\mathcal{L}$ is the set of all detected leaves
    \item $S_c(L_i)$ is the clutter/isolation score for leaf $i$
    \item $S_d(L_i)$ is the distance score for leaf $i$
    \item $S_v(L_i)$ is the visibility score for leaf $i$
    \item $w_c, w_d, w_v$ are the weights (typically 0.35, 0.35, 0.30)
\end{itemize}

\textbf{Leaf Selection:} Intuitively, the robot should target a leaf that is easy to access and likely to provide a good sample. We evaluate three main attributes for each candidate leaf: isolation, distance, and visibility. These correspond to how free the leaf is from clutter, how close it is to the camera (and robot tool), and how fully the leaf is observed. We quantify each attribute with a numeric score between 0 and 1, then combine them (with weights) to yield an overall selection score.

\begin{itemize}
    \item \textit{Clutter (Isolation) Score:} Using the 3D positions of leaves, we compute a signed distance field (SDF) in the horizontal plane that reflects the proximity of any point to the nearest leaf surface. A leaf receives a high clutter score if it is well separated from other leaves and surrounding structures. In practice, we take the leaf's 2D mask and compute the distance transform both inside and outside the mask. Points well inside the leaf (far from its edge) and with large outside clearance (far from other leaves) are ideal. We reward leaves that have an "optimal" border margin (around 20 px from edge in the image) by applying a Gaussian penalty if a leaf is too small or too large in extent. This favors medium-sized, isolated leaves. We also factor in the leaf's orientation: a leaf lying at an angle that separates it from neighbors (e.g., not overlapping directly above/below another) is scored higher.
    
    \item \textit{Distance Score:} Closer leaves are generally easier to reach and yield higher-resolution images. However, the robot's end-effector also has a minimum focus distance. We set an ideal distance range (around 30--50 cm from the camera) and score leaves based on how close they fall to this range. Using the depth to the leaf's centroid, we apply an exponential decay penalty beyond 0.5 m. Thus, leaves nearer than 0.5 m get close to 1.0 score, whereas very distant leaves (e.g., at 0.8 m) get significantly lower scores. This encourages the robot to pick leaves within comfortable reach of the gantry's Z-axis travel.
    
    \item \textit{Visibility Score:} This score reflects how completely the leaf is visible in the camera views. If a leaf is partially occluded by another (or by the edge of the camera frame), the segmentation mask will be incomplete or truncated. We estimate visibility by comparing the observed contour of the leaf to an expected full shape (via convex hull or fitted ellipse). Leaves with full contours and no significant occlusions get high visibility scores, whereas those with >20\% of their area obscured get penalized. This ensures the robot chooses a leaf it can see almost entirely, reducing uncertainty in shape and pose.
\end{itemize}

After computing these, we perform a weighted sum (35\% clutter, 35\% distance, 35\% visibility) to rank the leaves. This multi-criteria approach is akin to a Pareto optimization: the chosen leaf offers a good balance of being isolated, not too far, and clearly viewable. In our tests, this step often prefers an upper canopy leaf that sticks out from the rest – anecdotally, these were also the leaves a human scout would likely pick first.\\

\textbf{Grasp Point Selection:} Once a target leaf is selected, we determine where on that leaf to grasp. The grasp point must satisfy two conditions: the robot's end-effector can physically reach and engage the leaf at that point, and doing so will reliably yield a sample (i.e., puncture the leaf). We consider four factors for each candidate point on the leaf: flatness, approach angle, accessibility, and edge margin. Candidates are sampled across the leaf surface (we discretize the mask or evaluate on a grid of points inside the leaf).

The overall grasp point selection can be formulated as:

\begin{equation}
\begin{aligned}
G^* = \underset{p \in L^*}{\arg\max} \Big( &w_f F(p) + w_a A(p) \\
&+ w_e E(p) + w_{acc} Acc(p) \Big) \cdot (1 - S_{pen}(p))
\end{aligned}
\end{equation}

Where:
\begin{itemize}[label={--}]
    \item $G^*$ is the optimal grasp point
    \item $p$ is a candidate point on the selected leaf $L^*$
    \item $F(p)$ is the flatness score at point $p$
    \item $A(p)$ is the approach vector alignment score at point $p$
    \item $E(p)$ is the edge margin score at point $p$
    \item $Acc(p)$ is the accessibility score at point $p$
    \item $S_{pen}(p)$ is a penalty term for stem regions
    \item $w_f, w_a, w_e, w_{acc}$ are the weights (0.25, 0.40, 0.20, 0.15)
\end{itemize}

\begin{itemize}
    \item \textit{Flatness (Surface Smoothness):} The leaf should be locally flat at the grasp point so that the clamp can press evenly. Using the 3D point cloud of the leaf, we estimate the surface normal around each candidate (e.g., by fitting a plane to neighboring points). We then compute a "planarity" score: points where the local surface normal variance is low (the leaf is smooth, not deeply curved) get higher scores. This flatness score corresponds to the "leaf smoothness ($J_c$)" mentioned in our design. Typically, areas near the center of a leaf blade or along the main vein are flatter than edges that might curl.
    
    Mathematically, this is calculated as:
    \begin{equation}
    F(p) = \exp\Big(-\alpha \cdot \sqrt{|\nabla_x D(p)|^2 + |\nabla_y D(p)|^2}\Big)
    \end{equation}
    
    Where:
    \begin{itemize}[label={--}]
        \item $D(p)$ is the depth value at point $p$
        \item $\nabla_x D$ and $\nabla_y D$ are the gradients in x and y directions
        \item $\alpha$ is a scaling factor (typically 5.0)
    \end{itemize}
    
    \item \textit{Approach Vector Alignment:} Our robot is primarily limited to top-down approaches (due to the gantry structure). Thus, an ideal grasp point's surface normal should align closely with the vertical direction (camera's optical axis). We calculate the angle between the leaf's normal at a candidate point and the camera's $-Z$ axis. If this angle is small (leaf faces upward), the approach vector is near-vertical, which we favor. We convert this to a score (1.0 for perfectly upward-facing points, down to 0 if the leaf is oriented sideways). This factor (weighted highest at 40\%) strongly drives the selection towards points that the robot can approach perpendicularly, ensuring the microneedles will penetrate straight in. In some cases, the leaf's overall tilt is fixed; here the algorithm will choose a point along the midrib that maximizes the upward component.
    
    This alignment score is given by:
    \begin{equation}
    A(p) = \Big|\frac{\vec{v}(p) \cdot \vec{z}}{|\vec{v}(p)|}\Big|
    \end{equation}
    
    Where:
    \begin{itemize}[label={--}]
        \item $\vec{v}(p)$ is the vector from camera to point $p$
        \item $\vec{z}$ is the unit vector in the vertical direction (0,0,1)
    \end{itemize}
    
    \item \textit{Accessibility:} Even if a point is geometrically reachable, it might be at an extreme position relative to the robot or camera field of view (e.g., far to one side). We include an accessibility score that incorporates the distance of the point from the robot's origin and its position relative to the image center. Points roughly under the camera's central view and not at the boundary of the robot's range are preferred. This helps avoid selecting a point that, say, lies at the very edge of the camera frame where calibration error is higher and the robot would have to stretch near its limits to reach.
    
    The accessibility score is calculated as:
    \begin{equation}
    Acc(p) = 0.7 \cdot \Big(1 - \frac{d(p, c)}{d_{max}}\Big) + 0.3 \cdot \cos(\theta(p))
    \end{equation}
    
    Where:
    \begin{itemize}[label={--}]
        \item $d(p, c)$ is the distance from point $p$ to the camera center in image space
        \item $d_{max}$ is the maximum possible distance in the image
        \item $\theta(p)$ is the angle between the vector to point $p$ and the forward direction
    \end{itemize}
    
    \item \textit{Edge Margin:} To avoid tearing the leaf or missing it, the grasp point should not be too close to the leaf's edge. We ensure a margin by using the distance transform of the leaf mask – this gives the distance of each point to the leaf boundary. We assign zero score to points closer than 5 mm to the edge, and highest score to points well inside (e.g., $>$20 mm from edge). This corresponds to the "graspable area ($J_g$)" criterion: essentially the point should have enough leaf area around it to be securely gripped.
    
    The edge margin score is defined as:
    \begin{equation}
    E(p) = 
    \begin{cases}
    0, & \text{if } d_{edge}(p) < 5 \text{ mm} \\
    \frac{d_{edge}(p)}{d_{max}}, & \text{if } 5 \text{ mm} \leq d_{edge}(p) \leq 20 \text{ mm} \\
    1, & \text{if } d_{edge}(p) > 20 \text{ mm}
    \end{cases}
    \end{equation}
    
    Where:
    \begin{itemize}[label={--}]
        \item $d_{edge}(p)$ is the distance from point $p$ to the nearest edge of the leaf
        \item $d_{max}$ is set to 20 mm
    \end{itemize}
\end{itemize}

\begin{figure}[t]
    \centering
    \begin{minipage}{\columnwidth}
        \centering
        \includegraphics[width=0.95\columnwidth]{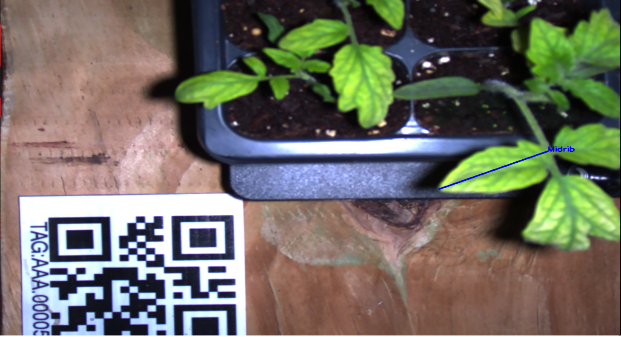}
        \vspace{-0.3cm}
        \caption*{(a) Raw input image from the stereo camera}
    \end{minipage}
    
    \vspace{0.5cm}
    
    \begin{minipage}{\columnwidth}
        \centering
        \includegraphics[width=0.95\columnwidth]{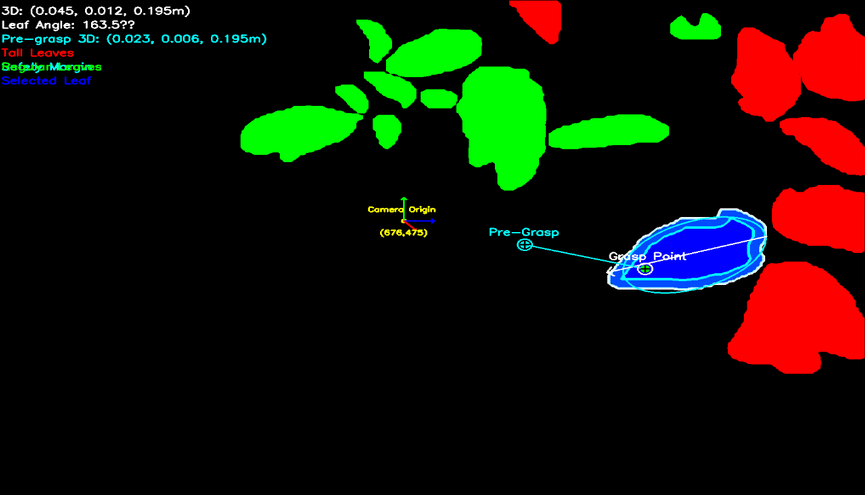}
        \vspace{-0.3cm}
        \caption*{(b) Traditional CV grasp planning output}
    \end{minipage}
    
    \caption{Grasp point selection visualization. (a) Raw camera image showing leafs and optimal leaf's midrib. (b) Output of the traditional computer vision pipeline with grasp point, pre-grasp pose, and selected leaf highlighted along with 3D coordinates and approach angle.}
    \label{fig:grasp-selection}
\end{figure}

Each candidate point on the target leaf gets a weighted sum of these factors. The point with the maximum combined score is chosen as the grasp location. Figure~\ref{fig:grasp-selection} illustrates an example: subfigure (a) shows the raw stereo image input, and subfigure (b) shows the output from our geometric CV pipeline with the chosen grasp and pre-grasp points highlighted in blue. Key annotations include the 3D coordinates of the grasp point and the leaf angle relative to the camera. This pipeline effectively encodes expert knowledge into geometric criteria, selecting interior, flat, accessible regions while avoiding tips and edges.

\subsection{Motion Planning and Reload Mechanism}

\textbf{Motion Planning:} Once a target leaf and grasp point are determined, the T-Rex robot must execute the maneuver to physically acquire the sample. The motion planning subsystem is responsible for moving the end-effector to the leaf without colliding with the plant, performing the sampling action, and then retreating safely. We formulate this planning problem as a sequence of waypoints: a pre-grasp approach pose, the leaf grasp pose, and a post-grasp retreat.

Our system leverages the Open Motion Planning Library (OMPL) through the MoveIt interface for trajectory generation. The planning pipeline can be expressed as:

\begin{equation}
\mathcal{P}(q_s, q_g, \mathcal{C}) \rightarrow \tau
\end{equation}

Where:
\begin{itemize}[label={--}]
    \item $\mathcal{P}$ is the planning algorithm (RRTConnect by default)
    \item $q_s$ is the start configuration
    \item $q_g$ is the goal configuration
    \item $\mathcal{C}$ is the configuration space with obstacles
    \item $\tau$ is the resulting trajectory $\{q_1, q_2, ..., q_n\}$
\end{itemize}

We configure our planners with specific parameters to ensure smooth, collision-free motion through the greenhouse environment. For the RRTConnect algorithm, we use a longest\_valid\_segment\_fraction of 0.01, which helps maintain precision when operating near plants. The planning time is set to 10 seconds with 10 planning attempts, providing multiple opportunities to find an optimal path.

For each grasp, a pre-grasp pose is defined a few centimeters above the selected point, aligned with the leaf's normal. The robot's inverse kinematics easily computes the joint angles for this pose since the gantry's prismatic axes can position the end-effector $(x,y,z)$ and the revolute joints orient the tool to match the normal. The robot then plans a straight-line trajectory from its current pose to this pre-grasp pose using a top-down approach strategy to minimize disturbance to surrounding leaves. This approach avoids any lateral sweeping through the canopy, reducing the risk of damaging plants or disturbing the target leaf before sampling.

Motion execution is handled by ROS controllers that generate smooth joint trajectories. We use trapezoidal velocity profiles with relatively low maximum velocity (20\% of maximum) and acceleration (20\% of maximum) values to gently move the heavy gantry axes, preventing oscillations that could disturb plant tissue. The ROS control system \cite{Macenski2022} proved sufficient for coordinating the multi-joint motion in our structured greenhouse environment.\\

\textbf{Microneedle Array Reload Mechanism:} A key innovation in the T-Rex system is the automated microneedle array reload mechanism that enables extended operation without human intervention. Since each leaf sampling operation consumes a single microneedle array, continuous operation requires a way to replace used arrays. Our reload subsystem addresses this challenge through a cartridge-based approach.

\begin{figure}[t]
    \centering
    \begin{minipage}{0.75\columnwidth}
        \centering
        \includegraphics[width=\textwidth]{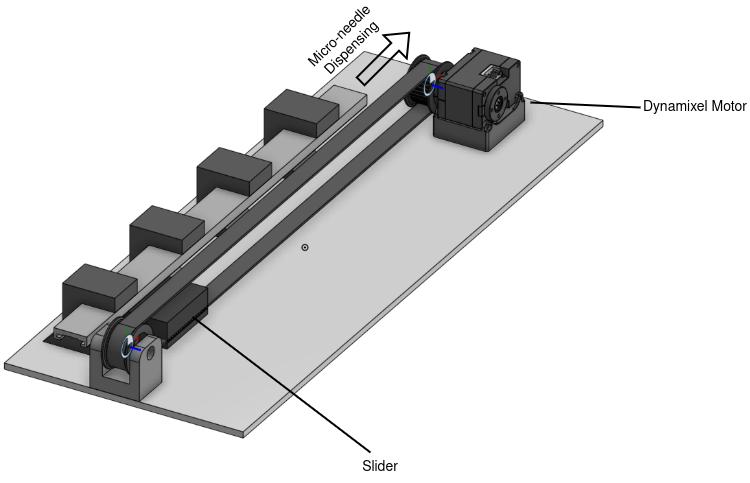}
        \vspace{-0.3cm}
        \caption*{(a) Microneedle array dispensing system}
    \end{minipage}
    
    \vspace{0.5cm}
    
    \begin{minipage}{0.48\columnwidth}
        \centering
        \includegraphics[width=\textwidth]{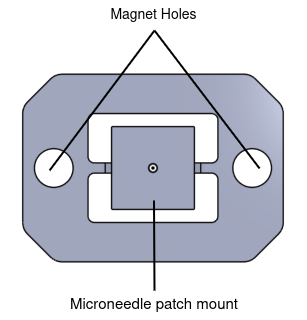}
        \vspace{-0.3cm}
        \caption*{(b) Close-up view of a single microneedle array}
    \end{minipage}
    
    \caption{The microneedle array reload system: (a) Linear actuator-driven dispensing mechanism that houses multiple microneedle arrays for sequential loading; (b) Individual microneedle array showing the sampling surface that makes contact with the leaf.}
    \label{fig:reload-mechanism}
\end{figure}

The reload mechanism consists of a Dynamixel motor-driven linear actuator that advances a cartridge containing multiple pre-loaded microneedle arrays (Figure~\ref{fig:reload-mechanism}). The system is controlled via a U2D2 communication board that interfaces with the main ROS control system. After each successful leaf sampling operation, the gantry navigates to the reload station positioned at a fixed location in the workspace.

The reload sequence follows a precise protocol:
\begin{enumerate}
    \item The end-effector approaches the reload station and aligns with the dispenser
    \item The linear actuator advances the cartridge, pushing a new microneedle array toward the end-effector
    \item The microneedle arrays feature an innovative magnetic attachment system with two magnetic holes on each array 
    \item The holes contain magnets with strategically arranged polarities: opposite poles at one end (attracting) and same poles at the other end (repelling)
    \item When the new array advances, it pushes against the used array in the end-effector, creating both attractive and repulsive magnetic forces
    \item This magnetic configuration causes the used array to be easily ejected while simultaneously guiding the new array into proper alignment and position
    \item The gantry retracts and is ready for the next sampling cycle
\end{enumerate}

This magnetic transfer mechanism ensures reliable array replacement without requiring precise force control or additional actuators, greatly simplifying the reload operation. The automation enhancement significantly increases the operational efficiency of the T-Rex system, allowing it to sample dozens of leaves consecutively without manual reloading. The current design accommodates 10 microneedle arrays per cartridge, though this capacity could be scaled based on application requirements.

Throughout this process, precision control is paramount. We calibrate the kinematics so that the targeting error is within a few millimeters. As a safeguard, if the vision system’s 3D estimate were significantly off (e.g., the leaf is actually a bit further than thought), the clamp might miss. However, our evaluations show the average grasp placement error to be $\sim$50 pixels in the image frame ($\sim$10 mm in world units), which is within the tolerance of the end-effector’s size. The ROS control system (citing Macenski \textit{et al.}, 2022 for ROS design \cite{Macenski2022}) proved sufficient for coordinating the multi-joint motion. We did not yet integrate advanced collision avoidance beyond the simple vertical approach; fortunately, by design, collision risk is minimal as long as the selected leaf was isolated (which our selection algorithm ensures). If an unexpected obstacle is encountered or a joint fails to reach commanded position, the system aborts and raises an error – an aspect to improve with more feedback sensors in future. Overall, the motion planning is relatively straightforward given the structured environment: essentially a pick-and-place movement where the “pick” is a leaf in situ. This simplicity is one advantage of focusing on greenhouse automation; it allowed us to script reliable motions without complex real-time path re-planning. The next section discusses the microfluidic DNA extraction accomplished by the end-effector after the leaf is grasped.

\begin{figure}[t]
    \centering
    \begin{minipage}{0.48\columnwidth}
        \centering
        \includegraphics[width=\textwidth]{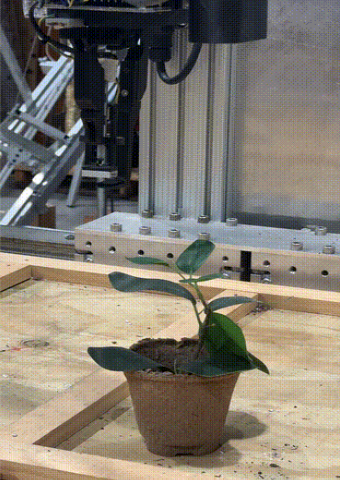}
        \vspace{-0.3cm}
        \caption*{(a) Start of grasp: approaching the leaf}
    \end{minipage}
    \hfill
    \begin{minipage}{0.48\columnwidth}
        \centering
        \includegraphics[width=\textwidth]{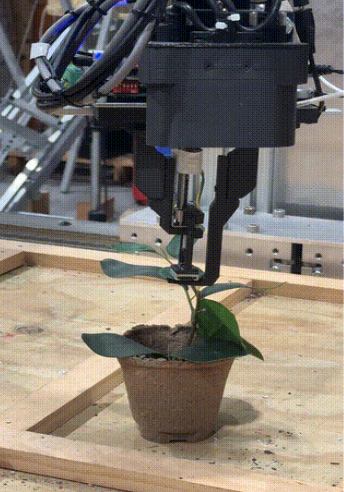}
        \vspace{-0.3cm}
        \caption*{(b) Grasp complete: microneedle fired}
    \end{minipage}
    
    \caption{Real-world grasp execution. (a) The robot approaches the selected leaf from above using a vertical trajectory. (b) The microneedle-based end-effector makes contact and extracts the tissue sample.}
    \label{fig:grasp-execution}
\end{figure}




\section{Results}

We evaluated the T-Rex system in a controlled lab environment using an artificial plant model to simulate tomato plant geometry. Instead of live plants, a plastic leaf replica was used to test grasping performance across diverse spatial configurations. The fake plant was manually repositioned between trials to present varying leaf orientations and locations relative to the camera and robot arm. This allowed systematic validation of the robot's ability to perceive, plan, and grasp under changing visual and geometric conditions without the variability of live foliage.

A total of 24 autonomous grasp-and-extract trials were conducted under these varying poses. In each trial, the robot captured stereo images, identified an optimal leaf, and attempted to execute the full sampling maneuver. Key performance metrics from these trials are summarized in Table~\ref{tab:performance-metrics}.\\

\begin{table}[t]
\centering
\caption{Key performance metrics across 24 autonomous grasp trials using the T-Rex system.}
\label{tab:performance-metrics}
\begin{tabular}{|p{4cm}|p{3cm}|}
\hline
\centering\textbf{Metric} & \centering\textbf{Value} \tabularnewline
\hline
Number of Trials & \centering 24 \tabularnewline
Grasp Success Rate & \centering 66.6\% \tabularnewline
Average Grasp Point Error & \centering $\sim$45 px (~10-12 mm) \tabularnewline
Failure Causes & \centering 25\% grasp point, 75\% approach angle\tabularnewline
Multi-Leaf Sampling Demo & \centering 5 plants in 6 minutes \tabularnewline
\hline
\end{tabular}
\end{table}

\textbf{Grasp Success Rate:} The primary metric is the percentage of trials in which the robot successfully clamped the target leaf without any mis-segmentation or execution failure. A trial was considered successful if the leaf was cleanly gripped by the end-effector and the vision system had correctly segmented the target without ambiguity. Mis-segmentation refers to cases where the YOLOv8 pipeline failed to properly identify and segment a leaf, particularly when leaves were highly tilted relative to the camera, preventing them from being processed by the grasp selection pipeline. Out of 24 trials, 16 met this criterion, resulting in a grasp success rate of approximately 66.6\%. The remaining 8 failures were due to either incorrect approach angles that caused the end-effector to miss or make only partial contact with the leaf due to poor grasp point selection. Figure~\ref{fig:grasp-execution} illustrates a successful execution with the robot approaching from the top and deploying the microneedle to engage the leaf.\\

\textbf{Accuracy of Targeting:} Since no formal ground-truth annotations were available for the fake plant trials, we qualitatively assessed accuracy through manual inspection. After each grasp attempt, we visually verified whether the end-effector landed near the intuitive center of the selected leaf. The most common failure cases occurred when the grasp point was poorly selected—causing the clamp to either miss or only graze the edge of the leaf. These errors were generally linked to noisy depth data or distorted segmentation masks. Overall, the grasp point selection was deemed reasonable in most trials, aligning well with human intuition even without precise measurement tools.\\

\textbf{Leaf Selection and Multi-Leaf Scenarios:} Our Pareto scoring for leaf selection was qualitatively validated by observing which leaf the robot chose in multi-plant scenes. It consistently selected an upper, exposed leaf when available, which mirrors human preference for easy picks. One limitation noted is that the algorithm sometimes prefers a slightly smaller isolated leaf over a larger but semi-overlapping one, even if the larger leaf might yield more DNA. Incorporating a direct "leaf size" term or task-specific criterion (e.g., sampling a particular part of the plant) could modulate this. Nonetheless, the selection strategy proved effective in avoiding obviously problematic leaves (e.g., heavily shaded bottom leaves).\\

\textbf{Integration Performance:} The full system integration was demonstrated in a continuous run where the robot autonomously sampled 5 leaves sequentially (one per plant). The ROS-based architecture ensured reliable coordination between subsystems. Vision data was synchronized with robot positioning, and each step properly triggered the next in sequence. We encountered occasional software crashes due to GPU memory issues when processing high-resolution images. In one case, a synchronization error caused the robot to move before the depth map was fully updated, resulting in a failed grasp. These issues were resolved with improved checking mechanisms. This experience underscores that robust integration presents challenges comparable to individual component functionality.

\begin{table}[t]
\centering
\caption{System Integration Performance Metrics}
\label{tab:integration-metrics}
\begin{tabular}{|p{4cm}|p{4cm}|}
\hline
\textbf{Metric} & \textbf{Value} \\
\hline
Total run time for 5 plants & 6 minutes \\
Stereo processing latency & 0.55 seconds \\
Segmentation latency & 0.15 seconds \\
Single leaf planning and execution & 50-60 seconds \\
Comparison to manual sampling & 3-6× faster than human operation \\
\hline
\end{tabular}
\end{table}

\section{Discussion}

The results validate the feasibility of using T-Rex for automated leaf sample collection under controlled conditions. A grasp success rate of 66.6\% across varied plant orientations is promising for a first physical prototype, especially given the reliance on classical computer vision methods rather than learned grasping policies. While testing was conducted on artificial plant models, the system consistently identified feasible grasp points and executed motions without major collisions. Most failures stemmed from suboptimal grasp point selection or depth inaccuracies, highlighting vision robustness as a key improvement area.

Our multi-criteria leaf selection algorithm using Pareto optimization proved effective in identifying accessible leaves that would be natural choices for human operators. The mathematical formulation of our grasp point scoring system allowed for transparent decision-making that could be traced and understood, avoiding the "black box" nature of some learning-based approaches. This transparency is particularly valuable in research applications where understanding why a specific leaf was chosen can impact experimental validity.

The motion planning implementation using the OMPL framework demonstrated reliable performance, with the RRTConnect algorithm generating collision-free paths even in the constrained environment around plant structures. The low velocity and acceleration parameters we selected proved critical for maintaining stability during the approach and grasp phases. Future improvements could include more advanced path planning that allows angled approaches for nearly vertical leaves, moving beyond our current top-down strategy.\\

The microneedle array reload mechanism represents a significant advancement for system autonomy. By enabling extended operation without human intervention, the T-Rex system can potentially survey large numbers of plants in succession. The magnetic transfer design proved both simple and reliable, avoiding the need for complex force control or precise positioning during the reload phase. In future iterations, increasing the cartridge capacity beyond the current 10 arrays would further extend autonomous operation time.

Vision system improvements remain a clear development path forward. Higher resolution cameras or more sophisticated segmentation models could reduce instances of mis-segmentation. The recently introduced Segment-Anything Model (SAM) might be explored for zero-shot leaf segmentation, though our current YOLOv8 implementation performed adequately in controlled settings. Adapting to different plant morphologies (like soybean trifoliate leaves versus tomato simple leaves) may require adjusting the grasp point criteria or implementing domain adaptation techniques.

The integration of the complete microfluidic DNA extraction system will be validated in upcoming greenhouse trials. Initial lab validation has confirmed successful DNA extraction from our samples, but we have yet to conduct end-to-end pathogen detection on-site. The next research phase will involve inducing controlled disease (e.g., viral infection) in test plants and using T-Rex to monitor infection progression through regular sampling. We anticipate that automated sampling will dramatically reduce labor requirements for such monitoring, potentially enabling daily sampling of entire research greenhouses—a scale impossible with manual methods.

While our current system operates in structured indoor environments, the underlying principles could extend to field robotics applications. A ground rover equipped with a similar vision and sampling system might perform leaf sampling in open fields, though navigation and stabilization for precise needle insertion on uneven terrain would present additional challenges.\\

In conclusion, the T-Rex robot demonstrates that autonomous leaf sample extraction is achievable by combining advanced perception with classical planning approaches. It represents a significant step toward fully automated crop health surveillance systems that integrate detection and intervention in a single platform.

\section{Conclusion}

This paper presented T-Rex, a robotic system capable of autonomously selecting and extracting plant leaf samples for DNA analysis. T-Rex integrates techniques from robotics, computer vision, and microfluidics to address a key bottleneck in precision agriculture – the laborious process of tissue sampling for pathogen detection. The system architecture combines an overhead gantry manipulator with a real-time 3D vision pipeline (YOLOv8 + RAFT-Stereo) and a mathematically formulated grasp optimization approach using Pareto principles. We demonstrated that this hybrid approach yields robust performance without requiring large training datasets for the grasping task.

The motion planning implementation using the OMPL framework with RRTConnect algorithm generated reliable collision-free paths even in the constrained plant environment. Our careful parameterization of velocity and acceleration profiles ensured stable operation during critical leaf interactions. The innovative microneedle array reload mechanism with its magnetic transfer design represents a significant advancement for system autonomy, enabling extended operation cycles without human intervention.

Experimental results showed approximately 66.6\% success in fully autonomous leaf sample extractions, a promising result for a first-of-its-kind prototype. The lessons learned point toward several improvements: incorporating more advanced vision to handle dense foliage, refining motion strategies for complex leaf geometries, and expanding the system's applicability to diverse crops. Nonetheless, our findings confirm that an automated system can relieve humans of repetitive sampling tasks and perform them with consistency.

The immediate next steps involve integrating on-board DNA detection to shorten the loop from sampling to diagnosis, and scaling up trials to stress-test T-Rex in a production greenhouse setting with hundreds of plants. In the long term, this approach could be extended to continuously monitor crops and even take automated mitigation actions, contributing to resilient and smart agriculture.

In summary, T-Rex validates the concept of autonomous plant sample collection, marrying state-of-the-art AI vision with pragmatic engineering solutions like our mathematical grasp selection model and efficient reload mechanism. It opens the door to new capabilities in agricultural robotics – where machines not only observe plants, but also physically interact with them to obtain biochemical insights. The work presented here lays a foundation upon which more sophisticated and generalizable plant interaction robots can be built.

\section{Acknowledgments}
The development of T-Rex was supported by the USDA-NIFA Cyber-Physical Systems program (Award \#2021-67021-34037) to AS and GK at CMU, and SL, BV and CL at Virginia Tech. The metagenomic analysis and machine learning components were supported by USDA FACT-CIN (Award \#2021-67021-34343) to SL and BV labs at Virginia Tech. Additional support was provided by the NSF AI Institute for Resilient Agriculture (AIIRA, Award \#2021-67021-35329). The authors thank the field robotics team at CMU for their technical support, and collaborators at Virginia Tech for guidance on plant pathogen assays. We would also like to thank Dexter Friis-Hecht, Kalinda Wagner, and Carolin Kiewel for their contributions to the computer vision pipeline, end-effector design, and circuit implementation.

\bibliography{aaai24}

\end{document}